# Building a language evolution tree based on word vector combination model

Zhu Gao[1]   Yanhui Jiang[2]   Junhui Gao[3,*]

1. School of Computer Science and Software Engineering, East China Normal University, Shanghai, China. 200062
10132510331@stu.ecnu.edu.cn
2. University of California, Berkeley, California, USA. CA94704
yanhuijiang1995@berkeley.edu
3. American and European International Study Center, Wuxi, Jiangsu, China. 214001
*. Corresponding Author, Email: jhgao68@163.com

**Abstract**

In this paper, we try to explore the evolution of language through case calculations.

First, we chose the novels of eleven British writers from 1400 to 2005 and found the corresponding works; Then, we use the natural language processing tool to construct the corresponding eleven corpora, and calculate the respective word vectors of 100 high-frequency words in eleven corpora; Next, for each corpus, we concatenate the 100 word vectors from beginning to end into one; Finally, we use the similarity comparison and hierarchical clustering method to generate the relationship tree between the combined eleven word vectors. This tree represents the relationship between eleven corpora. We found that in the tree generated by clustering, the distance between the corpus and the year corresponding to the corpus are basically the same. This means that we have discovered a specific language evolution tree.

To verify the stability and versatility of this method, we add three other themes: Dickens's eight works, the 19th century poets' works, and art criticism of recent 60 years. For these four themes, we tested different parameters such as the time span of the corpus, the time interval between the corpora, the dimension of the word vector, and the number of high-frequency public words. The results show that this is fairly stable and versatile.

## 1 Introduction
**Language evolution**

The question of language evolution has bothered human beings for a long time because of lack of empirical evidence. There are several general theories about language evolution: (1) Darwin's theory of biological language evolution. This view points out that the evolution of language is a gradual occurrence, the process of natural selection, the evolution of language is similar to the evolution of species; (2) Chomsky's view of language evolution. According to this view, the evolution of language comes from mutations in genes, and the evolution of language is a "great leap"[1][2]; (3) Pingke's view of language evolution. Pinker puts forward the language instinct that states that language is not only gifted but that language evolution is a long-term gradual process [3][4][5].

We have artifacts such as fossils to trace back studies to conclude the origin and evolution of the species in nature. However, language has no fossil to study which will certainly increase the difficulties of language evolution study.

**Natural language processing**

With the development of information technology and genomics, biologists have studied the evolution of genetic differences between different species. This inspired us to study language evolution by retaining the texts of hundreds or even thousands of years to date. In fact, more and more ancient and modern works covering history, literature, art, and politics have been turned into electronic documents. Even more gratifying is that with the development of natural language understanding technology, we can analyze the language more deeply.

Natural language processing (NLP) is an area of computer science and artificial intelligence concerned with the interactions between computers and human (natural) languages, in particular how to program computers to process and analyze large amounts of natural language data.

Accurately representing the distance between two documents has far-reaching applications in document retrieval [6], news categorization and clustering [7][8], song identification [9], and multilingual document matching [10].

In natural language, we can compare the differences between two or more documents in a variety of ways. For example, the literature [11] uses the word frequency change to analyze the 60-year trend of art criticism, the literature [12] [13] uses the TF-IDF method, and the literature [14] uses the word vector model to compare the changes in the United States over the past 100 years. The last paper used a new approach to quantitatively study historical trends using word vectors—here focused on the development of stereotypes in the United States in both gender and race in the 20th and 21st centuries.

Corpus refers to a large collection of well-sampled and processed electronic texts, on which language studies, theoretical or applied, can be conducted with the aid of computer tools.

The corpus can be composed of many documents. We have already mentioned the method of document comparison. How should the corpus be compared? Let us introduce the word vector model first.

**Word vector model and corpus comparison**

Probabilistic models such as Latent Dirichlet Allocation [15] are standard tools for analyzing text data. However, such models use bag-of-words representations. In contrast, word embeddings (real-valued vectors representing words in a vocabulary) were first introduced by [16] but popularized by Mikolovet al. [17] under the 'word2vec' moniker. The embedding space returned by a model trained on a sufficiently large and informative corpus models a notion of semantic similarity as cosine distance between word vectors.

There has been some work on comparing corpora based on word vector models, and we introduce two of them here.

Example 1 [18] is related to music. The paper applies a word embedding model to a large symbolic corpus of classical music to learn an embedding space where chords are represented by real-valued vectors. In early classical music, the first two principal components of the embed-dings of major triads form a circle. In music from later composers, this circular topology is less evident. The order in which major triads are arranged on this structure corresponds to their order in the circle of fifths. The emergence and perturbation of this structure is justified by reasoning about the probabilistic embedding model and stylistic trends in the composition of classical music. The paper shows how this technique is useful for large-scale, quantitative stylistic analysis of music, and musical document similarity in general, by using our learned embeddings and the word-mover's distance to classify composers.

Example 2 [19] is about comparison between poets. This paper tries to find out five poets' (Thomas Hardy, Wilde, Browning, Yeats, and Tagore) differences and similarities through analyzing their works on nineteenth Century by using natural language understanding technology and word vector model. Firstly, we collect enough poems from these five poets, build five corpora respectively, and calculate their high-frequency words, by using Natural Language Processing method. Then, based on the word vector model, we calculate the word vectors of the five poets' high-frequency words, and combine the word vectors of each poet into one vector. Finally, we analyze the similarity between the combined word vectors by using the hierarchical clustering method. The result shows that the poems of Hardy, Browning, and Wilde are similar; the poems of Tagore and Yeats are relatively close—but the gap between the two groups is relatively large. In addition, we evaluate the stability of our approach by altering the word vector dimension and try to analyze the results of clustering in a literary (poetic) perspective. Yeats and Tagore possessed a kind of mysticism poetics thought, while Hardy, Browning, and Wilde have the elements of realism combined with tragedy and comedy. The results are similar comparing to those we get from the word vector model.

**Building a tree of evolution**

The concept of phylogenetic tree first appeared in the field of biology.

The idea of a "tree of life" arose from ancient notions of a ladder-like progression from lower into higher forms of life (such as in the Great Chain of Being). Early representations of "branching" phylogenetic trees include a "paleontological chart" showing the geological relationships among plants and animals in the book Elementary Geology, by Edward Hitchcock [20].

Charles Darwin (1859) also produced one of the first illustrations and crucially popularized the notion of an evolutionary "tree" in his seminal book The Origin of Species. Over a century later, evolutionary biologists still use tree diagrams to depict evolution because such diagrams effectively convey the concept that speciation occurs through the adaptive and semi random splitting of lineages. Over time, species classification has become less static and more dynamic [21].

For the study of phylogenetic trees in the field of biology, reference can be made to the literature [22] [23].

**Research in this paper**

In a paper comparing five poets [19], we propose a technique for comparing corpora, that is, comparing corpora at different time points by word vector combination and hierarchical clustering. But we didn't realize that the corpus was related to time, or to evolution, so in this article, we tried to use this technique to study the evolution of language. We analyzed 11 English novels with a time span of more than 600 years and successfully constructed a language evolution tree.

## 2 Data Sources

We choose English to study the evolution of language. English is the most widely spoken language in the world. Another reason is that the NLTK we choose can handle English easily. If you switch to other languages, you will have trouble.

Then we choose novels, and the content of these famous articles is well preserved, which can better reflect the evolution of language.

Finally, considering the wide spread and purity of language, we have only chosen famous British writers.

We selected 11 works from 1400 to 2005, as shown in the table below. The electronic versions of these novels are mainly from the following websites: http://www.freeclassicebooks.com/

Table 2-1 Novels of English writers

| Year | Title | Author |
|---|---|---|
| 1400 | The Canterbury Tales | Chaucer |
| 1595 | Romeo and Juliet | Shakespeare |
| 1667 | Paradise Lost | Milton |
| 1719 | Robinson Crusoe | Defoe |

| 1749 | The History of Tom Jones | Fielding |
| 1813 | Pride and Prejudice | Jane Austin |
| 1847 | Dombey and Son | Dickens |
| 1885 | Mayor of Casterbridge | Thomas Hardy |
| 1932 | On Forsyte 'Change | John Galsworthy |
| 1960 | The Loneliness of the Long-Distance | Alan Sillitoe |
| 2005 | Never Let Me Go | Kazuo Ishiguro |

**3 Methods**

Natural Language Toolkit, referred to as NLTK, is a Natural Language Processing kit and a often used Python library in NLP, which was developed by Steven Bird and Edward Loper in the information science department at University of Pennsylvania.

For the 11 British novels selected from 600 years, we used NLTK to create 11 corpora and calculate the 100 most frequent words.

For each corpus, the word vector of the 100 high-frequency words is calculated by word2vec, and the word vector dimension of each word is 100, and then the 100-word vectors are connected back and forth to form a 10,000-dimensional vector.

Then, we use the cosine method to calculate the distance between the 11 vectors. The cosine similarity is derived from the cosine of the angle between two vectors in the vector space to measure the difference between the two individuals. The closer the cosine is to 1, the closer the angle is to zero, ie the close similarity between the two vectors. This is called "cosine similarity".

After we get the distance between the 11 novels, we subtract this value by 1, which we consider as the distance between the 11 novels.

Finally, we use hierarchical clustering to calculate the relationship between the 11 novels.

Hierarchical clustering is a general term for a class of algorithms that continuously merges clusters from bottom to top, or continuously separate clusters from top to bottom to form nested clusters. This level of class is represented by a "tree" [24]. The Agglomerative Clustering algorithm is a hierarchical clustering algorithm. The principle of the algorithm is very simple. In the beginning, all the data points themselves are clustered, and then the two clusters closest to each other are found to be combined into one, and the above steps are repeated until the preset number of clusters is reached.

We use Python 3.5.3, NLTK, gensim to implement the construction of corpus and the calculation of high frequency word vectors. We implement hierarchical clustering with hclust in R language. The code implementation version is R 3.5.0 for Windows.

**4 Results**

The resulting 600-year British novel evolution tree is shown in Figure 4-1.

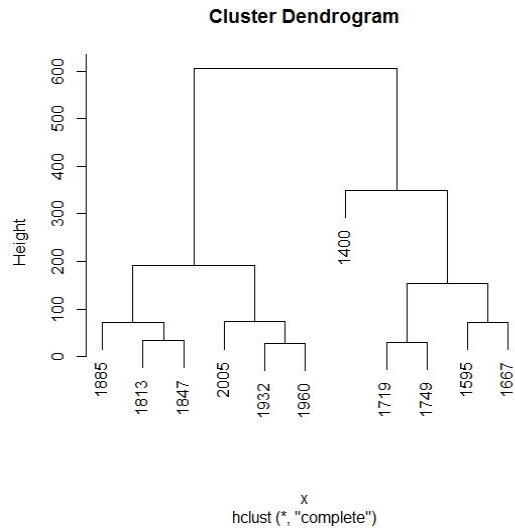

Figure 4-1 Evolution tree of 600-year British novel

Looking closely at Figure 4-1, we can see that the distance between the corpora is consistent with the distance between the years corresponding to the corpus. We have successfully constructed the evolutionary tree of English writers' novels. The 11 corpora can be divided into two groups, with a limit of 1800 years. You can get our code and data from GitHub: https://github.com/roygao94/LET.

**5 Discussion**

We have successfully constructed a phylogenetic tree of novels written by British writers for more than 600 years. Is this a method that can be promoted? The time span of the corpus, interval of text published in the corpus, the dimension of the word vector, and the number of high-frequency public words. Are these parameters a coincidence when constructing an art review phylogenetic tree? Let's discuss these issues separately.

**5.1 Corpus time span**

The theme of British writers' novels, our time span is 600 years. To study whether the time span affects the construction of the linguistic phylogenetic tree, we have added three themes, namely the poetry comparison of five poets in the 19th century, the art criticism of nearly 60 years, and the work of Dickens. The time spans of the four thematic corpora are 600 years, 129 years, 60 years, and 25 years.

First, continue to study the papers we have published about the comparison of five poets' poems [19].

Table 5-1 lists the birth and death times of the five poets, as well as other information that needs to be used later.

Table 5-1 80% of life time points

|  | Passed away | Born | 80% of life time points |
|---|---|---|---|
| Thomas Hardy | 1928 | 1840 | 1910 |
| Wilde | 1900 | 1854 | 1891 |
| Browning | 1889 | 1812 | 1874 |
| Yeats | 1939 | 1865 | 1924 |
| Tagore | 1941 | 1861 | 1925 |

From Table 5-1, we can calculate the time span of 129 years, the earliest is 1812, and the latest is 1941.

Considering that writers have generally published works in the age of infants and children, even in childhood, it is obviously not appropriate to set the average publication time at the midpoint of the life process. We chose the time point when they experienced 80% of life as the average publication time of the work and generated the corresponding evolution tree in the same way as before.

The result is shown in the following figure.

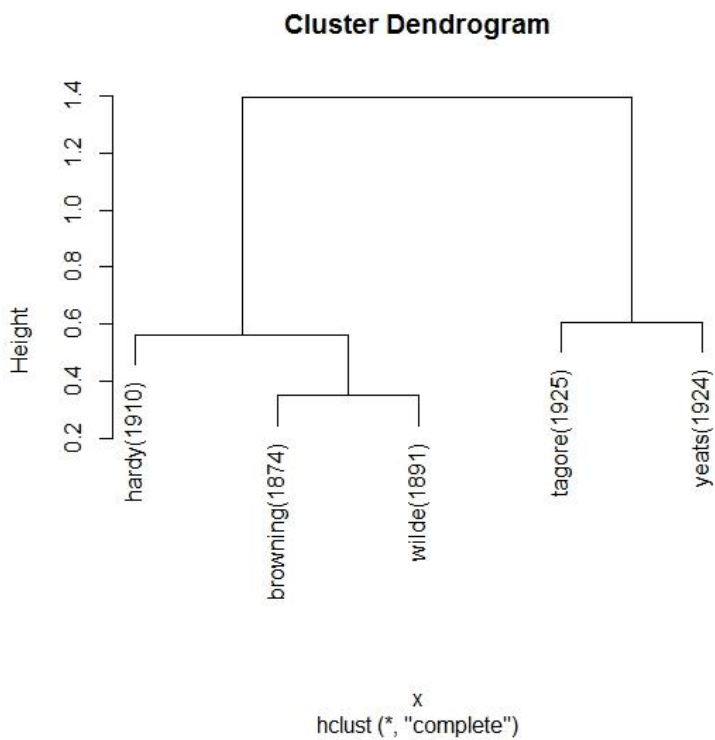

Figure 5-1 The evolutionary tree of the works of five poets

From the figure we can see that the distance between the corpus and the year corresponding to the corpus remains the same. In our previous papers, we tried to explain the tree from a literary perspective, but now we can explain it in terms of language evolution.

Next, continue to study the papers we have published about the 60-year art review [11]. In a similar way, the resulting 60-year art review evolutionary tree is shown in Figure 5-2.

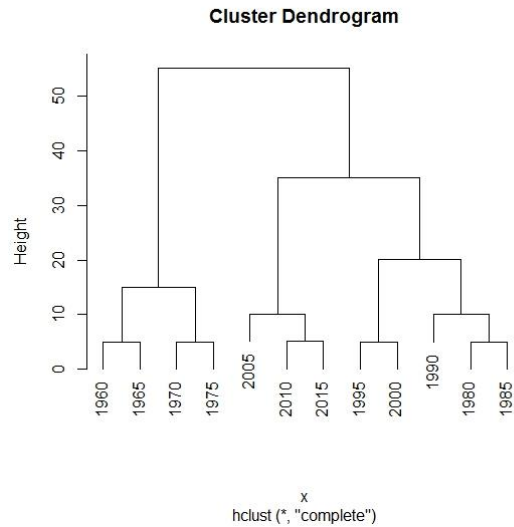

Figure 5-2 Evolutionary tree of 60 years of art criticism

Looking closely at Figure 5-2, we can see that the distance between the corpora is consistent with the distance between the years corresponding to the corpus. These 12 corpora can be divided into two groups, 1 group from 1960 to 1975, and group 1 from 1980 to 2015. The latter can be divided into two groups, 1980 to 2000 and 2005 to 2015.

Finally, we chose a topic that has a shorter time span than 60 years. We chose Dickens's eight novels published from 1836 to 1861. The time span was 25 years.

The theme and publication time of the novels are shown in Table 5-2. The electronic versions of these novels are from the following websites: http://www.freeclassicebooks.com/charles_dickens.htm。

Table 5-2 Dickens's 8 novels

|   | Year | Title |
| --- | --- | --- |
| 1 | 1836 | The Pickwick Papers |
| 2 | 1838 | Oliver Twist |
| 3 | 1843 | A Christmas Carol |
| 4 | 1847 | Dombey and Son |
| 5 | 1852 | Bleak House |
| 6 | 1854 | Hard Times |
| 7 | 1959 | A Tale Of Two Cities |
| 8 | 1861 | Great Expectations |

We generated the corresponding phylogenetic tree in the same way as before, and the result is shown in Figure 5-3. From the figure we can see that the distance between the corpus and the year corresponding to the corpus remains the same. The eight novels can be divided into two groups, 1836 to 1847, 1 group, and 1852 to 1861.

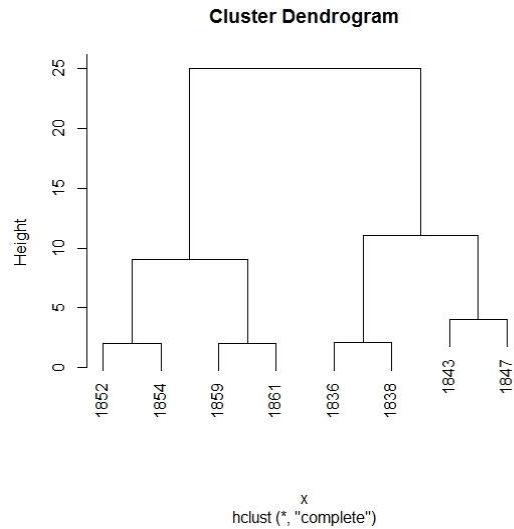

Figure 5-3 The evolutionary tree of Dickens's novel

## 5.2 Time interval between works within the corpus

We present the results of the calculation of the time interval between the four topic corpora in Table 5-3.

Table 5-3 Time interval between works

|  | time span | Time interval between works | Remarks |
|---|---|---|---|
| Works of 5 poets | 129 | 0.41 | 1/2.46 |
| Dickens | 25 | 3.33 | 25/8 |
| Art Reviews | 60 | 5.00 |  |
| British writer novel | 600 | 60 |  |

From Table 5-3, we know that the time intervals of the works in the four thematic corpora are 0.41, 3.33, 5 and 60 years respectively. The average number of works published by five poets per year is 2.46, and the calculation method is shown in Table 5-4.

Table 5-4 Number of poems per year

|  | Life expectancy | Number of poems | Number of poems per year |
|---|---|---|---|
| Thomas Hardy | 88 | 257 | 2.92 |
| Wilde | 46 | 96 | 2.09 |
| Browning | 77 | 63 | 0.82 |
| Yeats | 74 | 400 | 5.41 |
| Tagore | 80 | 86 | 1.08 |
| Average value |  |  | 2.46 |

## 5.3 Change in word vector dimension

In our published paper, we discussed this problem. We changed the 100-dimensional to 120-dimensional or 80-dimensional, and the relative position remained unchanged. Here we perform similar operations on the 60-year art review, and the results are shown in Figure 5-3.

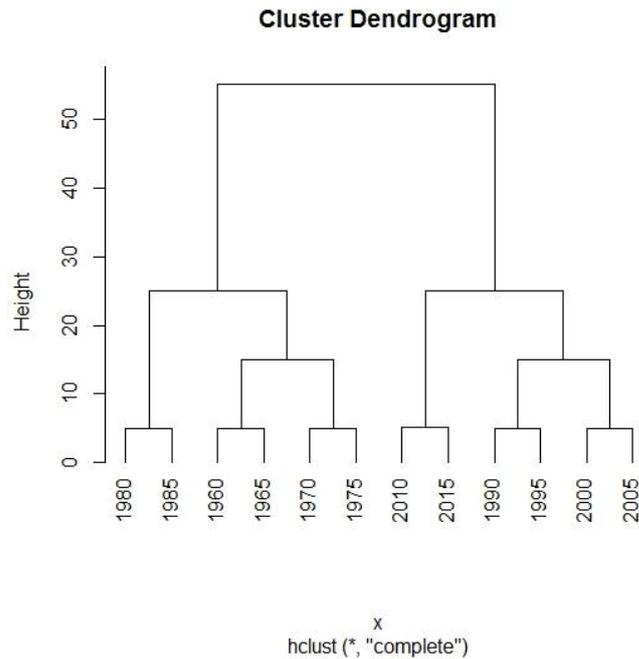

Figure 5-4 120-dimensional evolutionary tree

Figure 5-4 has some differences from Figure 5-2, but the consistency of time remains. We generate a new word vector model by changing the dimensions of the word vector and find that this consistency is quite stable.

**5.4 Number of high frequency words**

Here we discuss the effect of the number of high frequency words on the generation of phylogenetic trees. The high frequency words 100 and 10 are selected separately, and the generated similarity matrix is listed in Table 5-5 and Table 5-6.

Table 5-5 Similarity between corpora.
(the number of high frequency words is 100)

| id | 1836 | 1838 | 1843 | 1847 | 1852 | 1854 | 1859 | 1861 |
|---|---|---|---|---|---|---|---|---|
| 1836 | 0.000 | 0.375 | 0.405 | 0.299 | 0.286 | 0.294 | 0.326 | 0.348 |
| 1838 | 0.375 | 0.000 | 0.414 | 0.318 | 0.329 | 0.356 | 0.345 | 0.379 |
| 1843 | 0.405 | 0.414 | 0.000 | 0.443 | 0.348 | 0.313 | 0.360 | 0.346 |
| 1847 | 0.299 | 0.318 | 0.443 | 0.000 | 0.298 | 0.316 | 0.344 | 0.375 |
| 1852 | 0.286 | 0.329 | 0.348 | 0.298 | 0.000 | 0.240 | 0.223 | 0.299 |
| 1854 | 0.294 | 0.356 | 0.313 | 0.316 | 0.240 | 0.000 | 0.214 | 0.224 |
| 1859 | 0.326 | 0.345 | 0.360 | 0.344 | 0.223 | 0.214 | 0.000 | 0.256 |
| 1861 | 0.348 | 0.379 | 0.346 | 0.375 | 0.299 | 0.224 | 0.256 | 0.000 |

Table 5-6 Similarity between corpora.
(the number of high frequency words is 10)

| id | 1836 | 1838 | 1843 | 1847 | 1852 | 1854 | 1859 | 1861 |
|---|---|---|---|---|---|---|---|---|
| 1836 | 0.000 | 0.373 | 0.374 | 0.303 | 0.291 | 0.292 | 0.321 | 0.344 |
| 1838 | 0.373 | 0.000 | 0.400 | 0.320 | 0.326 | 0.352 | 0.340 | 0.373 |
| 1843 | 0.374 | 0.400 | 0.000 | 0.429 | 0.315 | 0.282 | 0.341 | 0.318 |
| 1847 | 0.303 | 0.320 | 0.429 | 0.000 | 0.309 | 0.331 | 0.351 | 0.381 |
| 1852 | 0.291 | 0.326 | 0.315 | 0.309 | 0.000 | 0.245 | 0.220 | 0.296 |
| 1854 | 0.292 | 0.352 | 0.282 | 0.331 | 0.245 | 0.000 | 0.206 | 0.221 |
| 1859 | 0.322 | 0.340 | 0.341 | 0.351 | 0.220 | 0.206 | 0.000 | 0.249 |
| 1861 | 0.344 | 0.373 | 0.318 | 0.381 | 0.296 | 0.221 | 0.249 | 0.000 |

The data of the above two tables is basically close in the corresponding position, and there is almost no difference in the phylogenetic tree drawn.

## 5.5 Inadequacies

There are some shortcomings in the study of text, such as the impact of the size of the corpus on the generation of phylogenetic trees.

## 6 Conclusion

In this paper, we use the word vector model, word vector combination and cluster analysis techniques to successfully construct four language phylogenetic trees. These evolutionary trees are very stable. The time span of the corpus, the time interval of text publication in the corpus, the number of word vectors, and the dimension of the word vector have not destroyed the time correctness of the evolutionary tree. The method proposed in this paper has certain versatility and can be extended to other fields than literature, such as constructing the evolutionary tree of music spectrum and constructing molecular evolution tree.

**References**
[1] Chomsky, N. (1995b). The Minimalist Program. Cambridge Mass: MIT Press.
[2] Chomsky, N. (2002). On Nature and Language. Cambridge: Cambridge University Press.
[3] Pinker, S. (1994). The Language Instinct. Great Britain: The Penguin Press.
[4] Pinker, S. (2004). Clarifying the logical problem of language acquisition. Journal of Child Language ,31.
[5] Pinker, S. (2006). The Blank Slate. General Psychologist, 41 (1).
[6] Salton, G. and Buckley, C. Term-weighting approaches in automatic text retrieval. Information processing & management, 24(5):513–523, 1988.
[7] Ontrup, J. and Ritter, H. Hyperbolic self-organizing maps for semantic navigation. In NIPS, volume 14, pp. 2001,2001.
[8] Greene, D. and Cunningham, P. Practical solutions to the problem of diagonal dominance in kernel document clustering. In ICML, pp. 377–384. ACM, 2006.
[9] Brochu, E. and Freitas, N. D. Name that song! In NIPS, pp. 1505–1512, 2002.
[10] Quadrianto, N., Song, L., and Smola, A. J. Kernelized sorting. In NIPS, pp. 1289–1296, 2009.
[11] Jiang, Y. and Gao, J. (2018) Word Frequency Analysis of American Contemporary Art Reviews from 1960 to

2015. Open Journal of Modern Linguistics, 8, 87-98. doi: 10.4236/ojml.2018.84010.

[12] Dumais, S & Platt, John & Heckerman, David & Sahami, M. (1998). Inductive learning algorithms and representations for text categorization. 148-155.

[13] JOACHIMST.A probabilistic analysis of the rocchio algorithm with TFIDF for text categorization Nashville：1997：143-151.

[14] Nikhil Garg，Londa Schiebinge，Dan Jurafsky，James Zou,2017.Word Embeddings Quantify 100 Years of Gender and Ethnic Stereotypes

[15] Blei, David M, Ng, Andrew Y, and Jordan, Michael I. Latent Dirichlet Allocation. Journal of Machine Learning Research, 3:993–1022, 2003. URL http://www.jmlr.org/papers/volume3/blei03a/blei03a.pdf.

[16] Bengio, Yoshua, Ducharme, Réjean, Vincent, Pascal, and Janvin, Christian. A Neural Probabilistic Language Model. The Journal of Machine Learning Research, 3:1137–1155, 2003. ISSN 15324435. doi: 10.1162/153244303322533223. URL http://jmlr.csail.mit.edu/papers/volume3/bengio03a/bengio03a.pdf.

[17] Mikolov, Tomas, Corrado, Greg, Chen, Kai, and Dean, Jeffrey. Efficient Estimation of Word Representations in Vector Space. ICLR, pp. 1–12, 2013. URL http://arxiv.org/pdf/1301.3781v3.pdf.

[18] Eamonn Bell, Jaan Altosaar. Applications of word embedding models to a classical music corpus: stylistic analysis and composer classification. Machine Learning for Music Discovery Workshop at the 33 rd International Conference on Machine Learning, New York, NY, 2016.

[19] Zhang, L. and Gao, J. (2017) A Comparative Study to Understanding about Poetics Based on Natural Language Processing. Open Journal of Modern Linguistics, 7, 229-237. doi: 10.4236/ojml.2017.75017.

[20] Elementary Geology, by Edward Hitchcock (first edition: 1840).

[21] https://en.wikipedia.org/wiki/Natural_language_processing

[22] N Jardine，CJV Rijsbergen，CJ Jardine. Evolutionary Rates and the Inference of Evolutionary Tree Forms.Nature, 1969, 224(5215):185-185.http://dx.doi.org/10.1038/224185a0

[23] Kishino, H. & Hasegawa, M. J Mol Evol (1989) 29: 170. https://doi.org/10.1007/BF02100115

[24] Theodoridis, S. and Koutroumbas, K. (2006) Pattern Recognition. 3rd Edition, Elsevier, Amsterdam.